%% file: Main.tex
\documentclass[letterpaper]{article} 
\usepackage{aaai25}  
\usepackage{times}  
\usepackage{helvet}  
\usepackage{courier}  
\usepackage[hyphens]{url}  
\usepackage{graphicx} 
\usepackage{natbib}  
\usepackage{caption} 
\usepackage{algorithm}
\usepackage{algorithmic}
\usepackage{multirow}
\usepackage[table]{xcolor}
\usepackage{subcaption}
\usepackage{color}
\usepackage{array}
\usepackage{booktabs}
\usepackage{colortbl}
\usepackage{amsmath}
\usepackage{bm}

\usepackage{enumitem}
\usepackage{amssymb}

\usepackage{newfloat}
\usepackage{listings}
\DeclareCaptionStyle{ruled}{labelfont=normalfont,labelsep=colon,strut=off} 
\floatstyle{ruled}
\newfloat{listing}{tb}{lst}{}
\floatname{listing}{Listing}
\title{Putting People in LLMs’ Shoes: \\ Generating Better Answers via Question Rewriter}
\author {
    Junhao Chen,
    Bowen Wang\thanks{Corresponding author.},
    Zhouqiang Jiang, 
    Yuta Nakashima
}
\affiliations {
    Osaka University, Japan\\
    \{junhao, zhouqiang\}@is.ids.osaka-u.ac.jp, \{wang, n-yuta\}@ids.osaka-u.ac.jp
}

\usepackage{bibentry}
\begin{document}

\maketitle

\begin{abstract}
Large Language Models (LLMs) have demonstrated significant capabilities, particularly in the domain of question answering (QA). However, their effectiveness in QA is often undermined by the vagueness of user questions. To address this issue, we introduce single-round instance-level prompt optimization, referred to as question rewriter.  By enhancing the intelligibility of human questions for black-box LLMs, our question rewriter improves the quality of generated answers. The rewriter is optimized using direct preference optimization based on feedback collected from automatic criteria for evaluating generated answers; therefore, its training does not require costly human annotations. The experiments across multiple black-box LLMs and long-form question answering (LFQA) datasets demonstrate the efficacy of our method. This paper provides a practical framework for training question rewriters and sets a precedent for future explorations in prompt optimization within LFQA tasks. Code is available at \url{https://github.com/3244we/Question-Rewriter}.
\end{abstract}

\input{Introduction}
\input{Related_Work}

\input{Method}
\input{Experiment}

\input{Analysis}

\input{Conclusions}

\input{Acknowledgments}

\appendix
\bibliography{aaai25}

\input{Ablation_Study_2}
\input{Dataset}

\input{Scoring_Template}

\end{document}

%% file: Introduction.tex
\section{Introduction}
Large language models (LLMs) have incorporated extensive world knowledge through learning vast publicly available corpora~\citep{DBLP:conf/emnlp/RobertsRS20}. It becomes increasingly common for people to seek knowledge from LLMs, especially in fields such as medicine and law~\citep{atallah2023large, harrington2023case, wang2024direct}. However, a near-paradoxical issue arises: \textit{People ask questions to get knowledge, while lack of knowledge often leads to poorly formulated or vague questions, hindering LLMs from providing precise answers}~\citep{kim-etal-2023-tree, DBLP:journals/corr/abs-2405-12063}. Fine-tuning can enhance LLMs' ability to understand vague questions, but most popular LLMs are black-box models, and their parameters are inaccessible. Thus, a step of transforming user questions into a format that LLMs can understand better, known as question rewriting, is crucial for question answering (QA).

Question rewriting is closely related to prompt optimization. A prompt is an input to LLMs that guides them in generating a specific response, including a question (possibly with some instructions), a conversation history, etc.~\cite{DBLP:journals/csur/LiuYFJHN23}. Question rewriting is prompt optimization solely for questions. Previous work on prompt optimization primarily focused on optimizing \textit{task-level} prompts. They decompose prompts into task-level instructions and \textit{instance-level} inputs, optimizing task-level instructions for better performance across all instances of the task~\citep{DBLP:journals/corr/abs-2309-16797, DBLP:journals/corr/abs-2309-08532, DBLP:journals/corr/abs-2401-08189}. 

\begin{figure}[t]
    \centering
    \includegraphics[width=1\columnwidth]{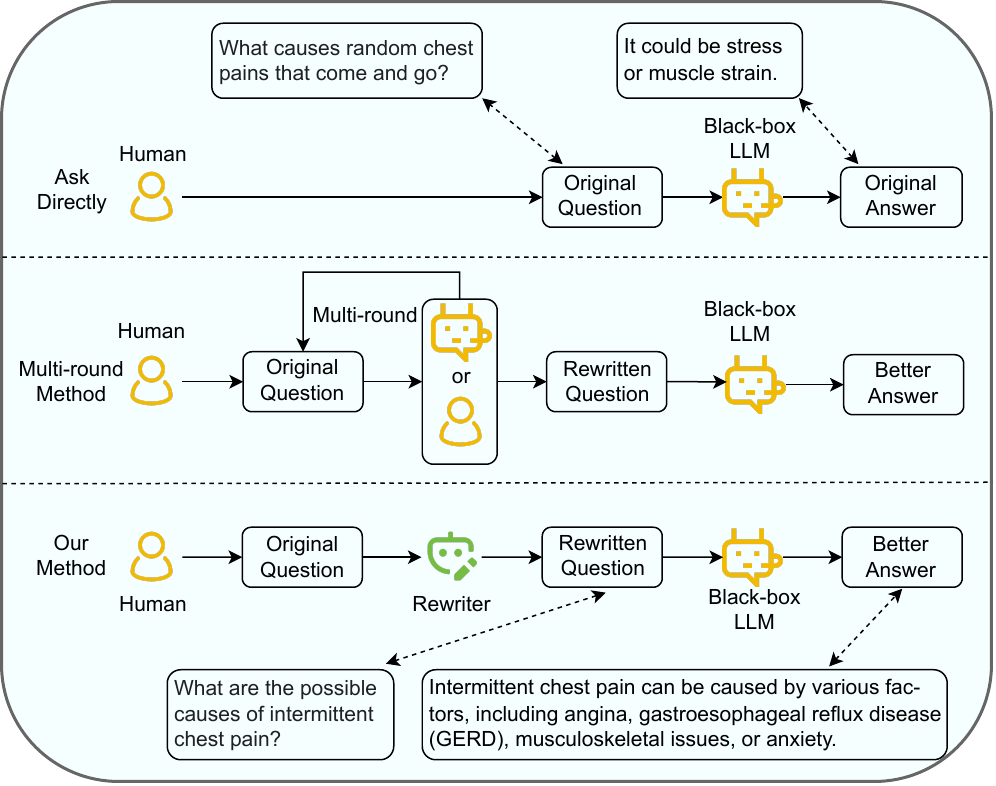}
    \caption{The original questions posed by the user are difficult for black-box LLMs to understand, resulting in poor answers. However, when the questions are rewritten by the rewriter, they become easier for LLMs to comprehend, leading to better answers.}
    \label{fig:proposal}
\end{figure}

Recent studies have shown that directly optimizing the prompts at the instance level offers more flexibility in prompt editing tailored for a given prompt\citep{DBLP:journals/corr/abs-2405-17346} and can lead to better responses \citep{DBLP:journals/corr/abs-2310-02107}. By obtaining feedback from humans or LLMs, they iteratively refine a given prompt, which requires multi-round interactions. In addition,  previous prompt optimization is mainly applied to arithmetic reasoning \cite{DBLP:journals/corr/abs-2110-14168} and short-form question answering (SFQA) \cite{DBLP:journals/tacl/KwiatkowskiPRCP19}, where the latter involves answers in few words. These tasks do not necessarily cover real-world QA scenarios. 


This paper proposes single-round instance-level prompt optimization, referred to as \textit{question rewriter}, aiming at optimizing questions for long-form question answering (LFQA), which is closer to real-world QA scenarios \citep{DBLP:journals/corr/abs-2309-08210}. The question rewriter serves as an intermediary between users and black-box LLMs to generate answers, as shown in Figure~\ref{fig:proposal}. When a user submits a question, our question rewriter scutches it to contextualizes the question for black-box LLMs to generate a more accurate answer.

The key to our method lies in obtaining supervising signals for optimizing the question rewriter. Different from arithmetic reasoning and SFQA, there is no unique best answer for LFQA questions; therefore, obtaining the optimal rewritten question as the ground truth to train a question rewriter is not trivial \citep{radford2018improving}. Our method, in contrast, assumes the presence of automatic criteria to evaluate generated answers, which are typically provided in LFQA datasets, and uses them to identify better and worse rewritten questions. With such supervising signals, we propose to use direct preference optimization (DPO) \citep{DBLP:conf/nips/RafailovSMMEF23} to train our question rewriter. Thanks to this design choice, our method does not necessitate costly human interactions used in reinforcement learning from human feedback \cite{DBLP:journals/corr/abs-2204-05862} and a differentiable reward model in proximal policy optimization \cite{DBLP:journals/corr/SchulmanWDRK17}.

\noindent\textbf{Contribution.} Our question rewriter is single-round prompt optimization without human interventions, which has not been explored so far. We experimentally show across multiple datasets and LLMs for answer generation that, with optimization by automatic evaluation criteria, the question rewriter can generate questions that end up with better answers. Intiguingly, our analysis implies that the question rewriter learns to generate non-leading and concise questions in a professional tone, which aligns with our intuitions when engineering a prompt. 


%% file: Related_Work.tex
\section{Related Work}
Early work for prompt optimization focused on white-box models 
\cite{DBLP:conf/emnlp/ShinRLWS20, DBLP:conf/emnlp/ShiHGHTZ23, DBLP:conf/acl/LiL20, DBLP:conf/emnlp/LesterAC21, DBLP:conf/naacl/ZhongFC21}. 
Due to the prevalent nature of black-box models such as GPT-3~\cite{DBLP:conf/iclr/PatelLRCRC23} and Claude~\cite{TheC3}, the following work targeted at these black-box models. Most works decomposed prompts into task-level (i.e., instructions) and instance-level (i.e., specific queries) and optimizing only task-level instructions. Some work assumed that input (i.e., text embeddings) and output (i.e., logits) are accessible and leveraged them to optimize prompts~\cite{DBLP:conf/icml/SunSQHQ22, DBLP:journals/corr/abs-2205-11200, DBLP:conf/emnlp/ChaiWSTW022}. Other recent work has attempted to remove this assumption. \citet{DBLP:conf/eacl/PrasadHZB23} and \citet{DBLP:conf/emnlp/PryzantI0L0023}  evaluate task-level instructions with small edits (e.g., replacing some phrases with their synonyms) and find better ones step by step. Evolutionary algorithms \cite{DBLP:journals/corr/abs-2309-16797, DBLP:journals/corr/abs-2309-08532}, reinforce learning \cite{DBLP:journals/tmlr/DiaoHXLLZZ23, DBLP:journals/corr/abs-2401-08189}, and planning-based methods \cite{DBLP:journals/corr/abs-2310-16427} have also been adopted. 

Some work fully utilized the inherent capabilities of LLMs to refine prompts. \citet{DBLP:conf/iclr/ZhouMHPPCB23} leverages an LLM to generate and refine the prompts iteratively, and \citet{DBLP:journals/corr/abs-2309-03409} provides the prompt optimization trajectory to an LLM, allowing for discovering inherent patterns and optimizing the prompt progressively. Other notable efforts, such as InstructZero~\cite{DBLP:journals/corr/abs-2306-03082} and INSTINCT~\cite{DBLP:journals/corr/abs-2310-02905}, have transformed black-box optimization into an iteretive optimization problem with white-box LLMs. 

All these works are task-level prompt optimization. Recent studies have shown that instance-level prompt optimization results in better performance by offering more specific prompts \citep{DBLP:journals/corr/abs-2405-17346,DBLP:journals/corr/abs-2310-02107}. These works iteratively refine a prompt at the instance level by obtaining feedback from humans or ChatGPT. We also adopt the instance-level approach, but unlike the other instance-level method, ours does not require feedback from LLMs or humans for multi-round optimization. We instead use preference optimization \citep{DBLP:conf/nips/RafailovSMMEF23} for our question rewriter that can optimize the questions at once without any feedback nor iterative refinement. We evaluate our method over LFQA tasks, whereas previous works mainly use arithmetic reasoning~\cite{DBLP:journals/corr/abs-2110-14168} and SFQA~\cite{DBLP:journals/tacl/KwiatkowskiPRCP19}. LFQA is closer to real-world QA scenarios \citep{DBLP:journals/corr/abs-2309-08210} and can highlight differences in the generated text.



%

%% file: Method.tex
\begin{figure}[t]
    \centering
    \includegraphics[width=0.5\textwidth]{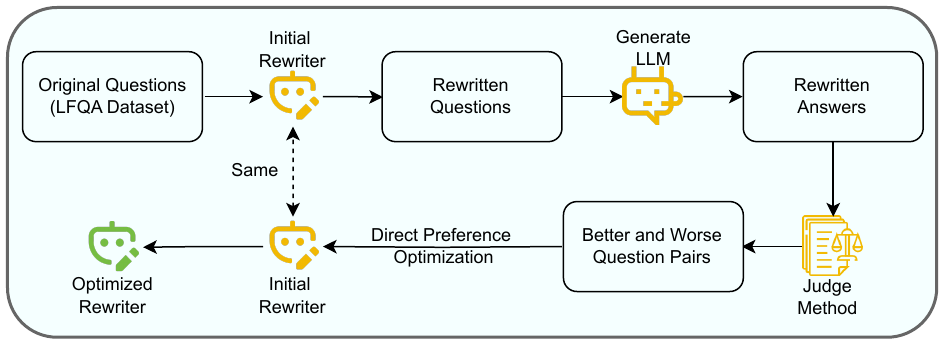}
    \caption{Pipeline of our method.}
    \label{fig:method}
\end{figure}

\section{Method}


Our question rewriter $R$ learns to rewrite questions so that an LLM can give a better answer for a rewritten question. We design our method under the assumption that the goodness of the answer to a certain question is automatically judgeable. With this assumption, we can sample rewritten questions and the corresponding answers using LLMs, contrasting them to learn desirable questions.

Figure \ref{fig:method} shows the pipeline of our method. Let $\mathcal{D} = \{(q, a)\}$ denote a training dataset of pairs of question $q$ and answer $a$, with an associated set $\mathcal{C} = \{c\}$ of automatic evaluation criteria $c$. Firstly, our pipeline rewrites questions for $q \in \mathcal{D}$. Then, $c \in \mathcal{C}$ evaluates the rewritten questions to make a set $\mathcal{P} = \{(\hat{q}, \check{q})\}$ of pairs of a better question $\hat{q}$ and a worse question $\check{q}$. Finally, we use DPO \citep{DBLP:conf/nips/RafailovSMMEF23} to train $R$ with $\mathcal{P}$.

\subsection{Sampling Rewritten Questions}

We use a pre-trained LLM $R_0$ to sample rewritten questions without fine-tuning as it offers sufficient capability for initial rewriting solely through prompt engineering. 
We use top-p sampling~\cite{radford2019language} to generate $K$ different rewritten questions $\mathcal{Q}(q) = \{r_{k}(q)|k=1,\dots,K\}$ of $q \in \mathcal{D}$, where $r_k(q)$ is the $k$-th rewritten question for $q$, with the predefined prompt $t$, i.e., $t$ equals:
\begin{quote}
    \texttt{\small Rewriting question to make it more understandable, just give me the rewritten question without any other word:}
\end{quote}
followed by $q$. 

\subsection{Making Better and Worse Question Pairs}\label{nrs}
Datasets for LFQA typically provide methods for evaluating generated answers. For instance, some datasets~\cite{DBLP:journals/corr/abs-2401-14493} are inspired by FActScore~\citep{DBLP:conf/emnlp/MinKLLYKIZH23} and annotate the facts required to answer each question, allowing LLMs to assess whether the corresponding facts are implied by or contradict the generated answers to derive scores for comprehensiveness and precision. Other datasets~\cite{DBLP:conf/acl/LinHE22} offer extensive binary annotations used to train classifiers to determine whether answers conform to certain attributes like truthfulness. Additionally, some datasets\footnote{\url{https://huggingface.co/datasets/tasksource/oasst1_pairwise_rlhf_reward}} are in the form of preference datasets, which provide pairs of samples, where one is better than the other. Such datasets can be used to train reward models to evaluate whether answers align with human preferences. We can use these automatic evaluation criteria as $\mathcal{C}$ to evaluate rewritten questions. Such automatic evaluation criteria substitute the human feedback typically used in previous methods \cite{DBLP:conf/nips/RafailovSMMEF23} to make $\mathcal{P}$. 

Let $L$ denote a pre-trained LLM for answer generation. For a question-answer pair $(q, a) \in \mathcal{D}$, we generate answers for all $q' \in \mathcal{Q}(q)$ as $a' = L(q')$. We also generate the answer to the original question $q$ as $\tilde{a} = L(q)$, which serves as the baseline to judge the goodness of rewritten questions. 

To make better-worse pairs, we first identify $q' \in \mathcal{Q}(q)$ that gives better answers and worse answers, collectively denoted by $\mathcal{Q}_+(q)$ and $\mathcal{Q}_-(q)$, respectively. Observing that criterion $c \in \mathcal{C}$ is often numerical\footnote{For example, $c(a')$ may be the probability of generated answer $a'$ is truthful.}, we judge $q'$ is better if $a'$ is larger than or equal to $\tilde{a}$ in terms of \textit{all} criteria and $a'$ is larger than $\tilde{a}$ at least one criterion,\footnote{It should be noted that a criterion may not always be \textit{larger-is-better}. For \textit{smaller-is-better} criteria, we just replace ``larger than'' with ``smaller than'' (and so in the inequalities). } i.e.,
\begin{align}
    \mathcal{Q}_+(q) = \{q' \in \mathcal{Q}(q)| \forall_{c \in \mathcal{C}} \; c(a') \geq c(\tilde{a}), \nonumber \\ \exists_{c \in \mathcal{C}} \: c(a') > c(\tilde{a})\}.
\end{align}
$\mathcal{Q}_-(q)$ is defined in the opposite way, i.e., $a'$ should be always worse than or equal to $\tilde{a}$ and $a'$ should be worse than $\tilde{a}$ for at least one criterion.



A better and worse question pair is created by picking one rewritten question from $\mathcal{Q}_+(q)$ and the other from $\mathcal{Q}_-(q)$. As we wish to train a model $R$ to generate good questions, we rank rewritten questions in $\mathcal{Q}_+$ according to a certain composition of all $c \in \mathcal{C}$,\footnote{For example, if all $c \in \mathcal{C}$ are a \textit{larger-is-better} type of criteria, the product of all criteria $\prod_c c(a')$ can be used. We design the composition for the set of criteria that a respective dataset provides.} and use the top $N_+$ questions. The set of chosen better questions is denoted by $\mathcal{Q}_+^\star(q)$. On the other hand, to avoid following DPO training only with easy negatives, we randomly choose $N_-$ questions in $\mathcal{Q}_-(q)$ and pair each of them with $\hat{q} \in \mathcal{Q}_+^\star(q)$. Formally, letting $\mathcal{S}$ denote randomly sampled $N_-$ questions from $\mathcal{Q}_-(q)$ without replacement, the set $\mathcal{P}(q)$ of better and worse question pairs for $q$ is given by:
\begin{align}
    \mathcal{P}(q) = \{(\hat{q}, \check{q})|\hat{q} \in \mathcal{Q}_+^\star(q), \check{q} \in \mathcal{S}(\mathcal{Q}_-(q))\}.
\end{align}
$\mathcal{P}(q)$ contains $N_+ \times N_-$ pairs when $|\mathcal{Q}_+(q)| \geq N_+$ and $|\mathcal{Q}_-(q)| \geq N_-$; otherwise, $|\mathcal{P}(q)|$ is smaller. The comparison of the different sampling combination\footnote{For instance,  $\mathcal{P}(q) = \{(\hat{q}, \check{q})|\hat{q} \in \mathcal{S}(\mathcal{Q}_+(q)), \check{q} \in \mathcal{Q}^\star_-(q)\}$, where $\mathcal{Q}^\star_-(q)$ denotes the set of bottom $N_-$ questions.} for $\mathcal{P}(q)$ can be found in Table \ref{tab:sample_methods}.


\subsection{Optimizing Question Rewriter}

Training our question rewriter $R$ is costly when it requires human feedback or a reward model that learns the human feedback. Fortunately, LFQA tasks typically offer automatic criteria to evaluate the goodness of generated answers. We can use the criteria to (indirectly) evaluate rewritten questions through evaluating their answers. 

Let $P_R(q'|t, q)$ denote the average probability of tokens in $q'$ given the predefined prompt $t$ and the original question $q$ with $R$, given by:
\begin{align}
    P_R(q') = \frac{1}{T} \sum_{k=1}^K p_R(w_k | t, q, w_{1:k-1}),
\end{align}
where $K$ is the length of $q'$; $p_R(w_t | t, q, w_{1:k-1})$ is the probability of token $w_t$ given $t$, $q$, and a set $w_{1:k-1}$ of tokens generated by the $(k-1)$-th step (i.e., $q' = w_{1:K}$).
$P_{R_0}(q')$ is defined likewise for the initial question rewriter $R_0$. DPO's training loss is given by:
\begin{equation}
\resizebox{210pt}{13.5pt}{$
L = - \mathbb{E}\left[\log \sigma \left( \beta \log \frac{P_R(\hat{q} )}{P_{R_0}(\hat{q})} - \beta \log \frac{P_R(\check{q})}{P_{R_0}(\check{q})} \right) \right]
$}
\end{equation}
where $\sigma$ is sigmoid, and $\beta$ is a hyperparameter that controls how much $R$ deviates from $R_0$, and the expectation is computed over $q \sim \mathcal{D}$ and $(\hat{q}, \check{q}) \sim \mathcal{P}(q)$. 

To mitigate the risk of overfitting, we use dropout in the model. Also, the original LFQA dataset is divided into three parts: training, validation, and testing. $R$ is trained on the training set (i.e., $\mathcal{D}$) for one epoch, and we select the best model that most prefer $\hat{q}$'s to $\check{q}$'s. Specifically, we define preference score PS as
\begin{equation}
\text{PS} = \mathbb{E}\left[ \mathbf{1}[P_R(\hat{q}|t, q) > P_R(\check{q}|t, q)]\right],
\end{equation}
where $\mathbf{1}[\cdot]$ gives $1$ if the given condition is satisfied, and otherwise $0$; the expectation is computed for all the $q$ from the validation set and $(\hat{q}, \check{q}) \sim \mathcal{P}(q)$.


%% file: Experiment.tex
\section{Experiments}


\subsection{Experimental Setup}

\paragraph{Dataset} We evaluate three distinct LFQA datasets, each equipped with automated evaluation criteria. 

\textbf{K-QA} \cite{DBLP:journals/corr/abs-2401-14493}, sourced from the medical domain, is designed to evaluate the factual comprehensiveness and precision of answers through metrics $S_{\text{comp}}$ and $S_{\text{cont}}$, employing a FActScore type method~\citep{DBLP:conf/emnlp/MinKLLYKIZH23}. To combine these two criteria for ranking rewritten questions in $\mathcal{Q}+(q)$, we first use $S_{\text{cont}}$ to rank them, and then use $S_{\text{comp}}$ if $S_{\text{cont}}$ is the same for multiple questions.

\textbf{TruthfulQA} \cite{DBLP:conf/acl/LinHE22}, covering multiple domains including health and law, assesses the truthfulness ($S_{\text{truth}}$) and informativeness ($S_{\text{info}}$) of answers. The evaluation criteria are implemented as binary classifiers. We use the probabilities for positive classes (\textit{truthful} for $S_{\text{truth}}$ and \textit{informative} for $S_{\text{info}}$). An overall score ($S_{\text{overall}}$) is computed as the product of these scores. For better rewritten pair ranking, we use $S_{\text{overall}}$.

\textbf{OASST1QA}, derived from the multi-turn dialogue alignment dataset OASST1\footnote{\url{https://huggingface.co/datasets/OpenAssistant/oasst1}}, incorporates a criterion $S_{\text{pref}}$ that measures human preference for answers using a pre-trained reward model. This dataset provides a single criterion for evaluation (i.e., $|\mathcal{C}| = 1$), so we directly use $S_{\text{pref}}$ for ranking better rewritten questions.

More details about these datasets and their evaluation criteria can be found in the appendix. Table \ref{tab:dataset_splits} summarizes the statistics on the datasets.

\begin{table}[t]
\centering
\renewcommand{\arraystretch}{1} 
\begin{tabular}{lcccc}
\toprule
\textbf{Dataset} & \textbf{Training} & \textbf{Validation} & \textbf{Testing} & \textbf{Total}\\
\midrule
\textbf{K-QA}        & 101  & 50   & 50   & 201\\
\textbf{TruthfulQA}  & 407  & 205  & 205  & 817\\
\textbf{OASST1QA}    & 1,000 & 93   & 93   & 1,186\\
\bottomrule
\end{tabular}
\caption{Statistics of LFQA datasets used to evaluate our method. Columns for \textbf{Training}, \textbf{Validation}, and \textbf{Testing} give the numbers of samples in respective dataset splits.}
\label{tab:dataset_splits}
\end{table}

\begin{table*}[ht]
\centering
\setlength{\tabcolsep}{10pt} 
\renewcommand{\arraystretch}{1} 
\begin{tabular}{llcccccc}
\toprule
& & \multicolumn{2}{c}{\textbf{K-QA}} & \multicolumn{3}{c}{\textbf{TruthfulQA}}        & \textbf{OASST1QA} \\
\cmidrule(lr){3-4} \cmidrule(lr){5-7} \cmidrule(lr){8-8}
\textbf{Model} & \textbf{Method}                                         & \bm{$S_\textbf{comp} \uparrow$}         & \bm{$S_\textbf{cont}\downarrow$}              & \bm{$S_\textbf{truth} \uparrow$} & \bm{$S_\textbf{info} \uparrow$} & \bm{$S_\textbf{overall} \uparrow$} & \bm{$S_\textbf{pref} \uparrow$}                      \\ \midrule
\multirow{4}{*}{\textbf{Llama-3-8B}}      & Original                        & 0.4573                   & 0.4400                  & 0.7683            & {\underline{0.9664}}     & 0.7397                    & 0.8654                             \\
                                          & Zero-Shot CoT                        & {\underline{0.4579}}              & \textbf{0.4000}         & 0.7476            & 0.9306           & 0.6938                    & {\underline{0.8838}}                        \\
                                          & Initial Rewriter            & 0.4262                   & 0.5000                  & {\underline{0.7914}}       & 0.9564           & {\underline{0.7566}}               & 0.8748                             \\
                                          & Ours                            & \textbf{0.4600}          & \textbf{0.4000}         & \textbf{0.8059}   & \textbf{0.9668}  & \textbf{0.7789}           & \textbf{0.9104}                    \\ \midrule
\multirow{4}{*}{\textbf{Mistral-7B-v0.2}}      & Original                        & 0.4374                   & \textbf{0.2200}         & 0.8364            & \textbf{0.9834}  & {\underline{0.8227}}               & 0.8281                             \\
                                          & Zero-Shot CoT                        & {\underline{0.4428}}              & 0.2800                  & {\underline{0.8423}}       & 0.9737           & 0.8199                    & \textbf{0.8908}                    \\
                                          & Initial Rewriter            & 0.4177                   & 0.3400                  & 0.7916            & 0.9689           & 0.7670                    & 0.8381                             \\
                                          & Ours                            & \textbf{0.4899}          & {\underline{0.2600}}             & \textbf{0.8474}   & {\underline{0.9788}}      & \textbf{0.8296}           & {\underline{0.8762}}                        \\ \midrule
\multirow{4}{*}{\textbf{Zephyr-7B-beta}}  & Original                        & 0.4396                   & 0.3400                  & {\underline{0.7644}}       & \textbf{0.9826}  & {\underline{0.7518}}               & 0.6369                             \\
                                          & Zero-Shot CoT                        & 0.4333                   & 0.3200                  & 0.7081            & 0.9705           & 0.6867                    & {\underline{0.7606}}                        \\
                                          & Initial Rewriter            & {\underline{0.4666}}          & \textbf{0.2200}         & 0.7353            & 0.9723           & 0.7167                    & 0.6417                             \\
                                          & Ours                            & \textbf{0.4702}          & {\underline{0.2600}}             & \textbf{0.7709}   & {\underline{0.9775}}      & \textbf{0.7528}           & \textbf{0.7768}                    \\ \midrule
\multirow{4}{*}{\textbf{Gemma-1.1-7B}} & Original                        & 0.4010                   & 0.5400                  & 0.6780            & \textbf{0.9716}  & 0.6554                    & 0.7428                             \\
                                          & Zero-Shot CoT                        & 0.4516                   & 0.5000                  & {\underline{0.7216}}       & 0.9454           & {\underline{0.6752}}               & {\underline{0.8632}}                             \\
                                          & Initial Rewriter            & {\underline{0.4928}}              & {\underline{0.4400}}             & 0.6415            & {\underline{0.9617}}      & 0.6124                    & 0.7955                             \\
                                          & Ours                            & \textbf{0.4956}          & \textbf{0.2200}         & \textbf{0.7224}   & 0.9558           & \textbf{0.6888}           & \textbf{0.9034}                    \\ \midrule
\multirow{4}{*}{\textbf{GPT-3.5-turbo}}   & Original                        & {\underline{0.4909}}              & 0.3200                  & {\underline{0.7451}}       & \textbf{0.9804}  & 0.7303                    & 0.7294                             \\
                                          & Zero-Shot CoT                        & 0.4748                   & \textbf{0.1600}         & 0.7413            & {\underline{0.9781}}      & {\underline{0.7237}}               & {\underline{0.8222}}                        \\
                                          & Initial Rewriter            & 0.4454                   & 0.3000                  & 0.7325            & 0.9768           & 0.7164                    & 0.7353                             \\
                                          & Ours                            & \textbf{0.4978}          & {\underline{0.2800}}             & \textbf{0.7574}   & 0.9682           & {\textbf{0.7309}}              & \textbf{0.8994}                    \\ \midrule
\multirow{4}{*}{\textbf{GPT-4o}}          & Original                        & 0.5167                   & 0.2800                  & {\underline{0.8812}}       & \textbf{0.9790}  & {\underline{0.8631}}               & {\underline{0.8532}}                       \\
                                          & Zero-Shot CoT                        & 0.4903                   & 0.3000                  & 0.8739            & 0.9611           & 0.8400                    & 0.8471                             \\
                                          & Initial Rewriter            & {\underline{0.5255}}             & \textbf{0.2400}         & 0.8593            & 0.9683           & 0.8329                    & 0.8461                             \\
                                          & Ours                            & \textbf{0.6253}          & \textbf{0.2400}         & \textbf{0.8880}   & {\underline{0.9722}}      & \textbf{0.8641}           & \textbf{0.9100}                    \\ 
\bottomrule
\end{tabular}
\caption{Comparison of different question rewriting methods across multiple datasets and LLMs for answer generation. We use automatic evaluation criteria associated with each dataset.  \textbf{Bold} indicates the best method, and \underline{underline} indicates the second best.}
\label{tab:main}
\end{table*}


\definecolor{mygreen}{rgb}{0.0, 0.5, 0.0} 
\definecolor{myred}{rgb}{0.5, 0.0, 0.0} 

\begin{table*}[htbp]
\small
\renewcommand{\arraystretch}{1} 
\centering
\begin{tabular}{@{}>{\centering\arraybackslash}m{0.18\linewidth}m{0.80\linewidth}@{}}
\toprule
\textbf{Original Question} & How do I know if it's a canker sore versus herpes? \\
\midrule
\textbf{Original Answer} 

\textcolor{mygreen}{$S_{comp}\uparrow: 2/6$} 

\textcolor{myred}{$S_{cont}\downarrow: 1$}& 
\textcolor{myred}{Canker sores and cold sores (caused by herpes simplex virus) are both types of oral lesions but they have distinct characteristics.}

\textbf{Canker Sore:}
\begin{itemize}
    \item \textcolor{mygreen}{\textbf{Location}: Canker sores typically occur on the inside of the mouth, tongue, lips or throat.}
\end{itemize}

\textbf{Cold Sore (Herpes Simplex Virus):}
\begin{itemize}
    \item \textcolor{mygreen}{\textbf{Location}: Cold sores typically occur on the lips, around the mouth, or on the nose.}   
\end{itemize} 

...
\\
\midrule
\textbf{Rewritten Question} & How can I distinguish between a canker sore and a herpes outbreak based on their symptoms and characteristics? \\
\midrule
\textbf{Rewritten Answer} 

\textcolor{mygreen}{$S_{comp}\uparrow: 3/6$} 

\textcolor{myred}{$S_{cont}\downarrow: 0$}& 

\textbf{Canker Sores (Aphthous Ulcers):}
\begin{itemize}
    \item \textcolor{mygreen}{\textbf{Location}: Typically occur on the inside of the mouth, tongue, lips, or throat.}
\end{itemize}

\textbf{Herpes Simplex Virus (HSV) Outbreaks:}
\begin{itemize}
    \item \textcolor{mygreen}{\textbf{Location}: Can occur anywhere on the body but most commonly on the mouth, lips, or genital area.}
    \item \textcolor{mygreen}{\textbf{Appearance}: Small, fluid-filled blisters that eventually crust over and form a scab.}
\end{itemize} 

...
\\
\bottomrule
\end{tabular}
\caption{An example of original and rewritten questions in K-QA with their answers. Text highlighted in green is consistent with the must-to-have fact, whereas red contradicts the must-to-have fact. Omitted text in answers, represented by ``...'', are irrelevant to any must-to-have fact.}
\label{tab:example}
\end{table*}

\paragraph{LLMs} The base model of our question rewriter $R$ (and $R_0$ is Llama3-8B-instruct, and the answer generation model $L$ is also Llama3-8B-instruct because it is one of the most powerful but small LLMs. $R$ is fine-tuned with our method, while $L$ is frozen.  Subsequently, we evaluate the generalizability of $R$ on multiple answer generation LLMs, including Llama3-8B-instruct\footnote{\url{https://huggingface.co/meta-llama/Meta-Llama-3-8B-Instruct}}, mistral-7B-instruct-v0.2\footnote{\url{https://huggingface.co/mistralai/Mistral-7B-Instruct-v0.2}}, zephyr-7B-beta\footnote{\url{https://huggingface.co/HuggingFaceH4/zephyr-7b-beta}}, gemma-1.1-7B-it\footnote{\url{https://huggingface.co/google/gemma-1.1-7b-it}}, gpt-3.5-turbo-1106, and gpt-4o-2024-05-13. They will be referred to as Llama3-8B, Mistral-7B-v0.2, Zephyr-7B-beta, Gemma-1.1-7B, GPT-3.5, and GPT-4o, respectively. It is worth noting that we only use $L$ as Llama3-8B-instruct to build $P$ for training $R$, and then test the generalizability of $R$ on other models.

\paragraph{Hyperparameters} We borrowed open-source code for DPO training over all three datasets\footnote{\url{https://github.com/eric-mitchell/direct-preference-optimization}}, which also provides the code for supervised fine-tuning of automatic criteria $S_\text{truth}$ and $S_\text{info}$ for TruthfulQA. During DPO training, we set the dropout rate to 0.8, the training batch size to 32, and the testing batch size to 64, maintaining all other parameters at their default settings in the source code. For sampling rewritten questions, we use top-p sampling, where the cumulative probability for top-p sampling is set to 0.999, and the temperature of $R_0$ is 1, to ensure diversity. We sample 100 unique rewritten questions for each of the original questions and terminate the sampling after 10,000 attempts. $N_+$ and $N_-$ are defaulted to (10, 20), (5, 10), and (4, 5) in K-QA, TQA, and OQA respectively. When multiplied by the number of samples in the corresponding training sets, they are around 20,000. The maximum token length is set to 512 during feedback collection and testing. During testing, to ensure reproducibility, we generate answers using greedy sampling.

\paragraph{Device} All our testing and training, except for the DPO training of OASST1QA, are conducted on a system equipped with four NVIDIA A100-PCIE-40GB. Due to the extensive length of OASST1QA, we only used samples whose question plus the prompt $t$ and rewritten questions $q'$ for question rewriting is less than or equal to 512 tokens and conducted the DPO training on a system with four NVIDIA A100-80GB-PCIe.

\paragraph{Baselines} In our experiments across different datasets and models, we compare our method with both the original questions and the initial Llama3-8B-instruct rewriter (without fine-tuning). To demonstrate the effectiveness of our approach, we also compare it with the widely used task-level prompting method, Zero-Shot Chain-of-Thought (Zero-Shot CoT) \cite{DBLP:conf/nips/KojimaGRMI22}. Other instance-level methods, such as PRoMPTed, require multiple rounds of interactions with humans or LLMs to obtain feedback and iteratively modify the prompt or question during inference, which is extremely costly in our QA scenarios. Therefore, we only perform comparisons with PRoMPTed and other question rewriting methods on the K-QA dataset and Llama3-8B-instruct.

\subsection{Result across models and datasets}

Table \ref{tab:main} summarizes our experimental results over three LFQA datasets. Our method demonstrates superior performance in most combinations of LLMs and datasets.

For the K-QA dataset, our method consistently shows the highest $S_\text{comp}$ scores across all models, especially with GPT-4o, where the improvement is most significant. Furthermore, it achieves the lowest $S_\text{cont}$ with half of the models and the second lowest in the rest. Notably, our method trained an effective question rewriter using only 151 samples (i.e., training set plus validation set), implying that our method requires only a small number of annotated samples in a real-world QA scenario. Table \ref{tab:example} shows an example from K-QA, in which $S_\text{comp}$ increases and $S_\text{cont}$ decreases after rewriting the original question.

For the TruthfulQA dataset, all methods generally reduce the informativeness (i.e., $S_\text{info}$) of the answers, and only ours gains the truthfulness score (i.e., $S_\text{truth}$). This is typical behavior in the TruthfulQA dataset as these two criteria exhibit a trade-off relationship \cite{DBLP:conf/acl/LinHE22}. Our method can increase $S_\text{overall}$, while the others reduce it. 

For OASST1QA, our method outperforms others except Mistral-7B-v0.2, where Zero-Shot CoT performs best.

\begin{figure}[htbp]
    \centering
    \includegraphics[width=1\columnwidth]{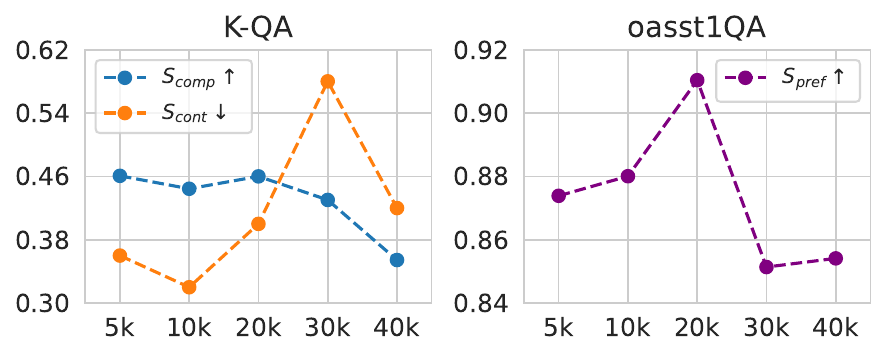} 
    \caption{Evaluating the impact of  $N_+$ and $N_-$ on the performance over K-QA and OASST1QA.}
    \label{fig:a}
\end{figure}

Overall, our method excels in all metrics and all datasets, not only on the Llama3-8B-instruct used for training $R$ but also on the other LLMs, demonstrating their generalizability to answer generation models. Thanks to this generalizability, the rewriter $R$ can be deployed without adjusting to individual (black-box) answer generation models. 

\subsection{Impact of $N_+$ and $N_-$}

The number of better and worse question pairs $\mathcal{P} = \cup_{q \in \mathcal{D}} \mathcal{P}(q)$ is determined by $N_+$, $N_-$, and $|\mathcal{D}|$ (i.e., $|\mathcal{P}| = N_+ \times N_- \times |\mathcal{D}|$), where the choice of $N_+$ and $N_-$ is rather arbitrary.  To explore the impact of $N_+$ and $N_-$ on the performance, we evaluated our method with varying $N_+$ and $N_-$ over K-QA OASST1QA with Llama-3-8B{\footnote{We evaluated $(N_+, N_-)$ in $\{(5, 10), (10, 10), (10, 20), (15,\\ 20), (20, 20)\}$ for K-QA and in $\{(1, 5), (2, 5), (4, 5), (5, 6), (5, \\8)\}$ for OASST1QA, which corresponds to $|\mathcal{P}|$ being 5k, 10k, 20k, 30k, and 40k, respectively.}}. The results are summarized in Figure~\ref{fig:a}.

For the K-QA dataset, finding the optimal values of $N_+$ and $N_-$ is not easy since multiple criteria are involved, but we would say $|\mathcal{P}|$ between 5k and 20k works well. Beyond 20k, $S_\text{comp}$ starts to decrease, and $S_\text{cont}$ spikes.  In OASST1QA, the performance increases along with $|\mathcal{P}|$ increases by 20k, then it decreases. The performance drops when $|\mathcal{P}|$ larger than 20k are attributed to overfitting during DPO training. These results highlight the necessity of adjusting $N_+$ and $N_-$ for each dataset. 

\subsection{Comparison with Other Methods}

To validate the superiority of our method, we compared it with other question rewriting methods on the K-QA dataset and Llama3-8B-instruct. In addition to the PRoMPTed method mentioned earlier, we considered two other closely related methods. The first method employs a transformer model to rewrite ill-formed questions into well-structured ones, while it does not evaluate the impact on answer quality \cite{chu2020ask}. We replicated this method using T5-Flan-base and achieved better performance on their original dataset. The second method, PRewrite \cite{DBLP:journals/corr/abs-2401-08189}, is an RL-based task-level prompt optimization method that uses RLHF, requiring a differentiable reward function. However, some LFQA datasets, such as K-QA, lack differentiable rewards. To address this, we implemented DPO, a variant of RLHF, to replicate it. We used prompts from it:

\begin{quote}
    \texttt{\small Rewrite the following instruction via rephrasing and/or adding specific requirements. Add instructions which would be helpful to solve the problem correctly. Output the new instruction only.}
\end{quote}
 to optimize the original task instruction: ``provide the answer:", and then appended the optimized instruction to the original questions to obtain answers. The optimized instruction can be found in the appendix.

\begin{table}[htbp]
\centering
\setlength{\tabcolsep}{1mm}
\renewcommand{\arraystretch}{1}
\begin{tabular}{@{}lcccccc@{}}
\toprule
\textbf{Metric} & \textbf{Orig.} & \textbf{Ours} & \textbf{PRewrite} & \textbf{T5} & \textbf{PRoMPTed} \\ \midrule
\bm{$S_\textbf{comp} \uparrow$} & 0.4573 & \textbf{0.4600} & 0.4409 & 0.4160 & \textbf{0.5012} \\
\bm{$S_\textbf{cont}\downarrow$} & 0.4400 & \textbf{0.4000} & \textbf{0.3600} & \textbf{0.3600} & 0.5400 \\ \bottomrule
\end{tabular}
\caption{Comparison of different question rewriting methods. \textbf{Bold} indicates the optimization method is better than the original results. \textbf{Orig.} indicates the original results. \textbf{T5} indicates the replicated result of ill-formed questions rewriting method}
\label{tab:ctl}
\end{table}

As shown in Table \ref{tab:ctl}, only our method demonstrates improvements in both metrics. While the PRoMPTed method improves $S_{comp}$, it significantly compromises $S_{cont}$, highlighting the superiority of our method in balancing these competing objectives. This further confirms that our method outperforms other question rewriting methods.

\subsection{Cross-Dataset Generalizability}

We explored the performance of rewriters trained on Llama3-8B-instruct across different datasets. Table \ref{tab:ad} shows that rewriters trained in the K-QA and TruthfulQA datasets can optimize the generated answers on the OASST1QA dataset, but each fails on one metric in their respective datasets. In contrast, rewriters trained on the OASST1QA dataset are almost ineffective on the other two datasets. This suggests that training on more complex LFQA datasets, which include multiple automatic evaluation criteria, yields rewriters with better generalizability.

\begin{table}[ht]
\centering
\setlength{\tabcolsep}{1mm} 
\renewcommand{\arraystretch}{1.1} 
\begin{tabular}{lcccccc}
\toprule
& \multicolumn{2}{c}{\textbf{K-QA}} & \multicolumn{3}{c}{\textbf{TruthfulQA}} & \textbf{OQA} \\
\cmidrule(lr){2-3} \cmidrule(lr){4-6} \cmidrule(lr){7-7}
\textbf{Rw.} & \bm{$S_\textbf{comp}$} & \bm{$S_\textbf{cont}$} & \bm{$S_\textbf{truth}$} & \bm{$S_\textbf{info}$} & \bm{$S_\textbf{overall}$} & \bm{$S_\textbf{pref}$} \\
\midrule
\textbf{Orig.}       & 0.4573 & 0.4400 & 0.7683 & 0.9664 & 0.7397 & 0.8654 \\
\textbf{Rw-K}     & \textbf{0.4600} & \textbf{0.4000} & \textbf{0.7834} & 0.9535 & \textbf{0.7454} & \textbf{0.8759} \\
\textbf{Rw-T}     & 0.4104 & \textbf{0.2800} & \textbf{0.8059} & \textbf{0.9668} & \textbf{0.7789} & \textbf{0.8839} \\
\textbf{Rw-O}     & 0.4510 & \textbf{0.4200} & 0.7622 & 0.9373 & 0.7155 & \textbf{0.9104} \\
\bottomrule
\end{tabular}
\caption{Performance of rewrites(Rw.) across datasets on Llama3-8B-instruct: OQA represents OASST1QA, Rw-K, Rw-T, and Rw-O represent rewriters trained on K-QA, TruthfulQA and OASST1QA, respectively. \textbf{Bold} indicates the rewriter performing better than the corresponding original for this LLM. Among all the metrics, except for \(\bm{S_\textbf{cont}}\), which is better when lower, all other metrics are better when higher.}
\label{tab:ad}
\end{table}

%% file: Analysis.tex
\section{Discussion}

Our question rewriters can significantly improve the representation of questions, making them more likely to obtain higher-quality answers with LLMs. To quantitatively analyze how attributes impact the evaluation criteria of generated answers, we study 50 original questions in KQA's test set and their rewritten versions, resulting in 100 questions in total. We adopt 10 attributes: non-leadingness, word choice, tone, conciseness, neutrality, grammar and spelling, structure, politeness, clarity, and emotion, which are identified by an LLM. For each attribute and each question, we use GPT-4o to assign a score ranging from 1 to 5 to the 100 questions. The definitions of attributes and the prompt templates are available in the extended version appendix.

To explore the important attributes that determine evaluation criteria $S_{\text{comp}}$ and $S_{\text{cont}}$, we use a random forest regressor that takes the attribute scores as input and predicts either $S_{\text{comp}}$ or $S_{\text{cont}}$ of each question. We train the regressors with these 100 questions and use the predictions again for them.\footnote{We consider the choice of using the test set for both training the regressors and analysis is reasonable as we wish to show how much the attribute scores can explain criteria $S_{\text{comp}}$ and $S_{\text{cont}}$.} The regressors yielded $R^2$ values of 0.56 and 0.55 for $S_{\text{comp}}$ and $S_{\text{cont}}$, respectively, demonstrating that the attribute scores are significantly correlated with the criteria. The random forest regressors provide feature importance, which, in our case, corresponds to the importance of each attribute. 

In addition to the attribute importance, we also examine whether each attribute has a positive or negative impact on the evaluation criteria. To this end, we define the impact by:
\begin{equation}
I_{la} = \hat{S}_{la} - \check{S}_{la}
\end{equation}
where $l \in \{\text{comp}, \text{cont}\}$ and $a$ is one of the 10 attributes; $\hat{S}_{la}$ and $\check{S}_{la}$ are the averages of evaluation criterion $S_l$ of questions whose attribute score for $a$ are among the top-50 and bottom-50, respectively. Specifically, $\hat{S}_{la}$ is given by:
\begin{align}
    \hat{S}_{la} = \frac{1}{50} \sum_{S_l \in \hat{\mathcal{S}}_{la}} S_l,
\end{align}
where $\hat{\mathcal{S}}_{la}$ is the set of scores $S_l$ of questions whose attribute scores are among top 50. $\check{S}_{la}$ is defined likewise. A higher $\hat{S}_{la}$, for instance, means that the attribute $a$ is positively correlated with $S_l$.

\begin{figure}[t]
    \centering
    
    \begin{subfigure}{0.495\columnwidth}
        \centering
        \includegraphics[width=\columnwidth]{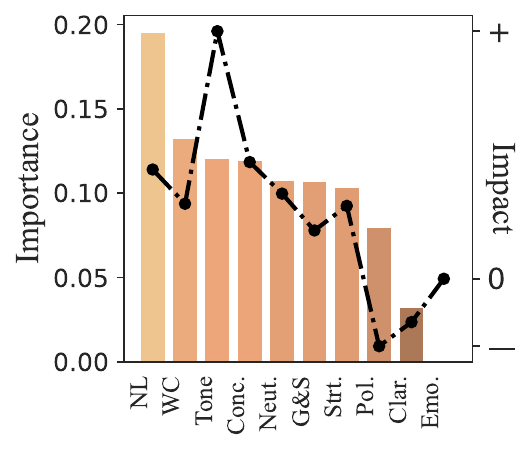}
        \caption{$S_{comp}$}
        \label{fig:entails}
    \end{subfigure}
    \hfill
    \begin{subfigure}{0.495\columnwidth}
        \centering
        \includegraphics[width=\columnwidth]{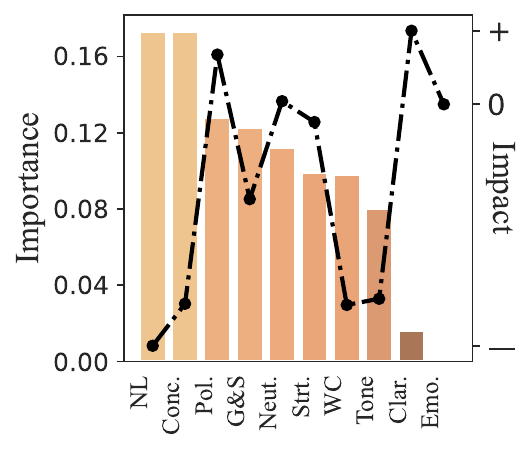}
        \caption{$S_{cont}$}
        \label{fig:contradicts}
    \end{subfigure}
    
    \caption{The importance and impact of attributes conciseness (Conc.), structure (Strt.), word choice (WC), emotion (Emo.), non-leadingness (NL), grammar and spelling (G\&S), neutrality (Neut.), tone, clarity (Clar.), and politeness (Pol.). The bar plots are important, while the line plots are impact.}
    \label{fig:ii}
\end{figure}

As shown in Figure \ref{fig:ii}, for $S_{\text{comp}}$, \textit{non-leadingness}, word choice, and tone are the most contributing attributes to regression, while for $S_\text{comp}$, \textit{non-leadingness}, \textit{conciseness}, and \textit{politeness} are important. Intriguingly, \textit{no-leadingness} for example, which means a question does not give some implication of a certain answer, is the most contributing attribute for both criteria. It is not straightforward to interpret this result, but being or not being leading may give some cues about attributes to be used for regression. At least, these attributes are somehow correlated with the evaluation criteria, and so the impact for these attributes can be meaningful.

As for the impact, we can see \textit{tone} gives a positive impact to $S_\text{comp}$, while \textit{non-leadingness} and \textit{conciseness} are negatively correlated with $S_\text{cont}$. A higher attribute score for \textit{tone} means the question is written in a formal language and a professional manner. We can reason that a formal language triggers expert knowledge encompassed in a LLM, which is likely to be written also in a formal language with proper wordings. Meanwhile, being non-leading and concise leads to lower $S_\text{cont}$, which is preferable. These results also make much sense; extra text in a question can lead to knowledge that is still relevant to the extra text but irrelevant to the question. Overall, our importance and impact analysis unveils that our question rewriter learns to generate professional, non-leading, and concise questions, which align with our intuitions, solely through supervision by $S_\text{comp}$ and $S_\text{cont}$.

%% file: Conclusions.tex
\section{Conclusions and Future Work}

This paper proposes single-round instance-level question optimization for LFQA tasks, coined \textit{question rewriter}. We employ DPO to optimize the question rewriter with automatic evaluation criteria. Our experimental results demonstrated that our question rewriter can generate questions that give better answers in terms of the automatic evaluation criteria. Meanwhile, although our method demonstrates some degree of cross-domain generalizability, it still has limitations in performance. Therefore, exploring completely domain-agnostic methods would be an interesting direction for future research.

%% file: Acknowledgments.tex
\section{Acknowledgments}
This work was supported by World Premier International Research Center Initiative (WPI), MEXT, Japan. This work is also supported by JST ACT-X Grant Number JPMJAX24C8 and JSPS KAKENHI No. 24K20795.

%% file: Ablation_Study_2.tex
\section{Comparison for Sampling Combination}

We test the performance of rewriters trained by $\mathcal{P}(q)$ sampled by different sampling combination on the K-QA dataset. There are four sampling combinations in total, where $\mathcal{Q}_+$ can be either the top $N_+$ questions or randomly sampled, and $\mathcal{Q}_-$ can be either the bottom $N_-$ questions or randomly sampled. Table \ref{tab:sample_methods} shows that rewriters trained by the Best-Random combination, which is the combination we currently use, perform the best, failing to optimize only one metric in one model. The Random-Random combination is effective across most models and metrics. The other three compositions all have 3 or 4 metrics worse than the corresponding original.

\begin{table*}[htbp]
\centering
\large
\setlength{\tabcolsep}{3pt} 
\renewcommand{\arraystretch}{1}
\begin{tabular}{@{}lrrrrrr@{}} 
\toprule
\textbf{Metric} & \textbf{Method} & \textbf{Original} & \textbf{Best-Random} & \textbf{Random-Worst} & \textbf{Best-Worst} & \textbf{Random-Random} \\
\midrule
\multirow{6}{*}{\bm{$S_\textbf{comp} \uparrow$}} & Llama-3-8B       & 0.4573 & \textbf{0.4600} & 0.4276          & 0.4026          & 0.4564 \\
                                                  & Mistral-7B-v0.2 & 0.4374 & \textbf{0.4899} & \textbf{0.4643} & 0.4304          & \textbf{0.4746} \\
                                                  & Zephyr-7B-beta  & 0.4396 & \textbf{0.4702} & \textbf{0.4607} & \textbf{0.4586} & \textbf{0.4650} \\
                                                  & Gemma-1.1-7B    & 0.4010 & \textbf{0.4956} & \textbf{0.4304} & \textbf{0.4065} & \textbf{0.4655} \\
                                                  & GPT-3.5-turbo   & 0.4909 & \textbf{0.4978} &     0.4905      & 0.4598          & \textbf{0.5302} \\
                                                  & GPT-4o          & 0.5167 & \textbf{0.6253} & \textbf{0.5759} & \textbf{0.5382} & \textbf{0.5520} \\
\midrule
\multirow{6}{*}{\bm{$S_\textbf{cont}\downarrow$}} & Llama-3-8B      & 0.4400 & \textbf{0.4000} & \textbf{0.4000} & \textbf{0.4000} & \textbf{0.4000} \\
                                                  & Mistral-7B-v0.2 & 0.2200 & 0.2600          & 0.2200          & 0.2400          & 0.3400 \\
                                                  & Zephyr-7B-beta  & 0.3400 & \textbf{0.2600} & \textbf{0.2800} & \textbf{0.3000} & \textbf{0.2600} \\
                                                  & Gemma-1.1-7B    & 0.5400 & \textbf{0.2200} & \textbf{0.3800} & \textbf{0.4000} & \textbf{0.3400} \\
                                                  & GPT-3.5-turbo   & 0.3200 & \textbf{0.2800} & \textbf{0.3000} & \textbf{0.3000} & \textbf{0.2600} \\
                                                  & GPT-4o          & 0.2800 & \textbf{0.2400} & \textbf{0.2000} & \textbf{0.2200} & 0.3200 \\
\bottomrule
\end{tabular}
\caption{Ablation study of the different sampling combination for $\mathcal{P}(q)$. \textbf{Bold} indicates the rewriter trained by the $\mathcal{P}(q)$ sampled by this combination performing better than the corresponding original for this LLM.}
\label{tab:sample_methods}
\end{table*}

%% file: Dataset.tex
\section{Dataset}

We conduct experiments on three LFQA datasets that have automated evaluation criterias.

\subsection{K-QA} 

A medical LFQA dataset~\cite{DBLP:journals/corr/abs-2401-14493} collected from a medical platform featuring questions from real-world patients in which questions are carefully curated to be ansared independently without additional patient history. The evaluation method for this dataset is inspired by the FActScore~\citep{DBLP:conf/emnlp/MinKLLYKIZH23}. Thus, in this dataset, each question is associated with must-have facts collected by professional annotators. These facts are then used to evaluate two metrics \(S_{\text{comp}}\) and )\(S_{\text{cont}}\), considering the comprehensiveness and  of the answers, respectively, defined as follows:

\[
\begin{aligned}
S_{\text{comp}}(a) &= \frac{\left| \left\{ x \in \text{Must\_Have} \mid a \text{ entails } x \right\} \right|}{\left| \text{Must\_Have} \right|} \\
S_{\text{cont}}(a) &= \left| \left\{ x \in \text{Must\_Have} \mid a \text{ contradicts } x \right\} \right|, 
\end{aligned}
\]

where \text{Must\_Have} represents the must-have facts for each question, \text{a entails x} indicates that the answer entails the corresponding fact, and \text{a contradicts x} indicates that the answer contradicts the corresponding fact. The entailment and contradiction are determined by LLM. The original study used GPT-4 for automatic evaluation. Due to time and cost constraints, we chose to use Llama-3-8B-instruct for automatic evaluation. The average accuracy of judgments across the dataset is  70\%.

\subsection{TruthfulQA} 

An LFQA data~\cite{DBLP:conf/acl/LinHE22} set is created to automatically assess the truthfulness and informativeness of the answers generated by LLMs. It spans 38 categories, including health, law, finance, and politics. In order to train the automatic evaluators, it includes truthful, untruthful, informatives, and uninformative answers for each question. After fine-tuning LLMs by specific templates, we can get automatic evaluators that can assess the truthfulness and informativeness of new answers to questions in this dataset. Upon inputting a question and an answer, the evaluators output a "yes" or "no", which is used to calculate the scores for truthfulness or informativeness using the following formula:

\[
S = \frac{p_{\text{yes}}}{p_{\text{yes}} + p_{\text{no}}}
\]

The original research use a GPT-3 base evaluation model. However, due to the deprecation of the corresponding API, we have transitioned to using Llama-3-8B-instruct as the base model. The batch size we use is 32, and the learning rate we use is $5 \times 10^{-6}$. To ensure the evaluator evaluates according to the original research standards with an accuracy exceeding 90\%, we conduct training for 2 epochs on truthfulness and 1 epoch on informativeness. We achieve accuracies of 93\% and 94\%, respectively. To prevent any potential information leakage during the evaluation of constructing the preference dataset, extra automatic evaluators are fine-tuned exclusively on the training set question and used to evaluate the metrics for answers to rewritten questions corresponding to the training set questions.

\subsection{OASST1QA} 

An LFQA dataset derived from the multi-turn dialogue alignment dataset, \textit{OASST1}\footnote{\url{https://huggingface.co/datasets/OpenAssistant/oasst1}}. OASST1 has several different responses along with human-annotated rankings in each dialog context. The dataset \textit{oasst1\_pairwise}\footnote{\url{https://huggingface.co/datasets/tasksource/oasst1_pairwise_rlhf_reward}} selects the highest and lowest ranked responses from each dialogue context to construct a preference dataset. The pretrained reward model\footnote{\url{https://huggingface.co/vincentmin/llama-2-7B-reward-oasst1}} we use is a binary classification model fine-tuned by oasst1\_pairwise. It can output the probability of whether a response is the highest-ranked response in a given dialogue context and the probability can be regarded as a measure of human preference, where a higher score indicates a greater preference.

We randomly sample 1000 single-round English QA pairs from the training set of \textit{oasst1\_pairwise} to serve as our training dataset. We use half of the single-round English dialog from the validation set of \textit{oasst1\_pairwise} as our preference dataset's validation set and the remaining half for final evaluation. We remeasure the discriminator, and its accuracy in the new testset is 75\%.

\subsection{Statistic} 

In both KQA and TruthfulQA, approximately 50\% of the questions are randomly selected as the training set, with 25\% of the questions serving as the validation set and test set. The selection method for OASST1QA is as described previously. Detailed statistics are in Table \ref{tab:dataset_splits}.


\section{Optimized Task Instruction of K-QA}

The optimized task instruction of K-QA obtained after migrating our method to task level is as follows:

\begin{quote}
    \texttt{\small $\ast$$\ast$To Solve the Problem Correctly:$\ast$$\ast$$\backslash$n$\backslash$nTo provide the correct answer, please follow these steps:$\backslash$n$\backslash$n1.$\ast$$\ast$Read the question carefully$\ast$$\ast$: Make sure you understand the question and its requirements. Take a moment to re-read the question to ensure you grasp the context and what is being asked.$\backslash$n2.$\ast$$\ast$Use the provided information$\ast$$\ast$: Refer to the given data, charts, or graphs to help you answer the question. Make sure to accurately interpret the information and use it to support your answer.$\backslash$n3.$\ast$$\ast$Calculate or derive the answer$\ast$$\ast$: Use mathematical formulas, calculations, or logical deductions to arrive at the correct answer. Double-check your work to ensure accuracy and precision.$\backslash$n4.$\ast$$\ast$Write your answer clearly and concisely$\ast$$\ast$: Use clear and concise language to express your answer. Avoid using jargon or technical terms unless absolutely necessary. Make sure your answer is easy to read and understand.$\backslash$n5.$\ast$$\ast$Check your work$\ast$$\ast$: Verify your answer by re-checking your calculations, formulas, or logical deductions. Ensure that your answer is accurate, complete, and relevant to the question.$\backslash$n$\backslash$n$\ast$$\ast$Additional Tips:$\ast$$\ast$$\backslash$n$\backslash$n$\ast$ Use a calculator or spreadsheet to help with calculations, if necessary.$\backslash$n$\ast$ Break down complex problems into smaller, manageable parts to make them easier to solve.$\backslash$n$\ast$ Use visual aids, such as diagrams or flowcharts, to help illustrate your answer.$\backslash$n$\ast$ Check for errors or inconsistencies in your work to ensure accuracy and precision.$\backslash$n$\backslash$nBy following these steps and tips, you will be well on your way to providing the correct answer and solving the problem correctly.}
\end{quote}

%% file: Scoring_Template.tex
\section{Question Evaluation Templates}
In this section, we present the templates used to evaluate question attributes. There are a total of ten attributes assessed, namely non-leading, word choice, tone, conciseness, neutrality, grammar and spelling, structure, politeness, clarity, and emotion. We provide detailed temples in Table \ref{tab:at1}, Table \ref{tab:at2} and Table \ref{tab:at3}.

\onecolumn
\begin{table}[t]
\small
\centering
\begin{tabular}{@{}>{\centering\arraybackslash}m{0.18\linewidth}>{\centering\arraybackslash}m{0.80\linewidth}@{}}
\toprule
\textbf{Attributes} & \textbf{Templates}\\
\midrule
\textbf{Word Choice} & Evaluate the word\ choice of the following question based on the specified criteria and provide an explanation for your rating.$\backslash$n$\backslash$n$\backslash$n- $\ast$$\ast$1 point - Very Poor:$\ast$$\ast$ The question uses non-professional or colloquial language.$\backslash$n- $\ast$$\ast$2 points - Poor:$\ast$$\ast$ The question uses some non-professional language or terminology.$\backslash$n- $\ast$$\ast$3 points - Moderate:$\ast$$\ast$ The question uses mostly professional language but may include some informal terms.$\backslash$n- $\ast$$\ast$4 points - Good:$\ast$$\ast$ The question uses professional and precise language.$\backslash$n- $\ast$$\ast$5 points - Excellent:$\ast$$\ast$ The question uses highly professional and precise language, appropriate for the context.$\backslash$n$\backslash$n$\backslash$n$\ast$$\ast$Question:$\ast$$\ast$ GIVEN QUESTION$\backslash$n$\backslash$n$\ast$$\ast$Rating:$\ast$$\ast$ \{1-5points\}$\backslash$n$\backslash$n$\ast$$\ast$Reasoning:$\ast$$\ast$$\backslash$n$\backslash$n$\ast$$\ast$Example:$\ast$$\ast$$\backslash$n$\backslash$n$\ast$$\ast$Question:$\ast$$\ast$ Can you please provide detailed information on the oral medication options available for treating scabies?$\backslash$n$\backslash$n$\ast$$\ast$Rating:$\ast$$\ast$ 5 points$\backslash$n$\backslash$n$\ast$$\ast$Reasoning:$\ast$$\ast$$\backslash$n- The word choice is highly professional and precise, suitable for a medical inquiry, making it appropriate and respectful for the context.\\\midrule
\textbf{Tone}  & Rate the tone of the following question based on the specified criteria and provide an explanation for your rating.$\backslash$n$\backslash$n$\backslash$n- $\ast$$\ast$1 point - Very Unprofessional Tone:$\ast$$\ast$ The question uses informal language, including slang or overly casual expressions that are inappropriate for a professional setting.$\backslash$n  $\backslash$n- $\ast$$\ast$2 points - Unprofessional Tone:$\ast$$\ast$ The question is somewhat informal, possibly including casual language that may not be suitable for all professional contexts.$\backslash$n  $\backslash$n- $\ast$$\ast$3 points - Moderately Professional Tone:$\ast$$\ast$ The question uses generally professional language but may include slight informal elements that could be polished further.$\backslash$n  $\backslash$n- $\ast$$\ast$4 points - Quite Professional Tone:$\ast$$\ast$ The question maintains a professional tone, using appropriate language for the context and avoiding casual terms.$\backslash$n  $\backslash$n- $\ast$$\ast$5 points - Very Professional Tone:$\ast$$\ast$ The question exemplifies a highly professional tone, using formal language that is perfectly suited for any professional setting.$\backslash$n$\backslash$n$\backslash$n$\ast$$\ast$Question:$\ast$$\ast$ GIVEN QUESTION$\backslash$n$\backslash$n$\ast$$\ast$Rating:$\ast$$\ast$ \{1-5 points\}$\backslash$n$\backslash$n$\ast$$\ast$Reasoning:$\ast$$\ast$$\backslash$n$\backslash$n$\ast$$\ast$Example:$\ast$$\ast$$\backslash$n$\backslash$n$\ast$$\ast$Question:$\ast$$\ast$ Can you please provide detailed information on the oral medication options available for treating scabies?$\backslash$n$\backslash$n$\ast$$\ast$Rating:$\ast$$\ast$ 5 points$\backslash$n$\backslash$n$\ast$$\ast$Reasoning:$\ast$$\ast$$\backslash$n- The question maintains a highly professional tone, using formal and respectful language suitable for a medical inquiry. The use of "please" adds a courteous touch, enhancing the professional quality of the communication. \\\midrule
\textbf{Conciseness}  &Evaluate the conciseness of the following question based on the specified criteria and provide an explanation for your rating.$\backslash$n$\backslash$n$\backslash$n- $\ast$$\ast$1 point - Very Verbose:$\ast$$\ast$ The question is verbose, containing much irrelevant information that obscures the main point.$\backslash$n- $\ast$$\ast$2 points - Verbose:$\ast$$\ast$ The question includes unnecessary details that detract from the main point, making it less concise.$\backslash$n- $\ast$$\ast$3 points - Moderately Concise:$\ast$$\ast$ The question is generally concise, though it contains a few extraneous details.$\backslash$n- $\ast$$\ast$4 points - Quite Concise:$\ast$$\ast$ The question is concise and to the point, with no unnecessary information.$\backslash$n- $\ast$$\ast$5 points - Extremely Concise:$\ast$$\ast$ The question is extremely concise, containing only essential information needed to answer the question effectively.$\backslash$n$\backslash$n$\backslash$n$\ast$$\ast$Question:$\ast$$\ast$ GIVEN QUESTION$\backslash$n$\backslash$n$\ast$$\ast$Rating:$\ast$$\ast$ \{1-5 points\}$\backslash$n$\backslash$n$\ast$$\ast$Reasoning:$\ast$$\ast$$\backslash$n$\backslash$n$\ast$$\ast$Example:$\ast$$\ast$$\backslash$n$\backslash$n$\ast$$\ast$Question:$\ast$$\ast$ Can you please provide detailed information on the oral medication options available for treating scabies?$\backslash$n$\backslash$n$\ast$$\ast$Rating:$\ast$$\ast$ 5 points$\backslash$n$\backslash$n$\ast$$\ast$Reasoning:$\ast$$\ast$$\backslash$n- The question is extremely concise, focusing solely on the essential information required. It directly asks for detailed options for oral medication without adding any superfluous details, ensuring the query is clear and straightforward.$\backslash$n$\backslash$nPlease rate the given question using this template. \\\midrule
\textbf{Non-leading}  &Evaluate whether the following question is non-leading based on the specified criteria and provide an explanation for your rating.$\backslash$n$\backslash$n$\backslash$n- $\ast$$\ast$1 point - Highly Leading:$\ast$$\ast$ The question is leading, suggesting a particular answer or outcome.$\backslash$n- $\ast$$\ast$2 points - Somewhat Leading:$\ast$$\ast$ The question is somewhat leading, with language that hints at a specific answer.$\backslash$n- $\ast$$\ast$3 points - Moderately Non-leading:$\ast$$\ast$ The question is mostly non-leading, with only slight suggestions towards an answer.$\backslash$n- $\ast$$\ast$4 points - Quite Non-leading:$\ast$$\ast$ The question is non-leading, asking for information without suggesting an answer.$\backslash$n- $\ast$$\ast$5 points - Completely Non-leading:$\ast$$\ast$ The question is entirely non-leading, purely seeking information without any implied bias towards certain answers.$\backslash$n$\backslash$n$\backslash$n$\ast$$\ast$Question:$\ast$$\ast$ GIVEN QUESTION$\backslash$n$\backslash$n$\ast$$\ast$Rating:$\ast$$\ast$ \{1-5 points\}$\backslash$n$\backslash$n$\ast$$\ast$Reasoning:$\ast$$\ast$$\backslash$n$\backslash$n$\ast$$\ast$Example:$\ast$$\ast$$\backslash$n$\backslash$n$\ast$$\ast$Question:$\ast$$\ast$ Can you please provide detailed information on the oral medication options available for treating scabies?$\backslash$n$\backslash$n$\ast$$\ast$Rating:$\ast$$\ast$ 5 points$\backslash$n$\backslash$n$\ast$$\ast$Reasoning:$\ast$$\ast$$\backslash$n- The question does not suggest any particular answer or bias, simply requesting information, making it completely non-leading.$\backslash$n$\backslash$nPlease rate the given question using these templates. \\
\bottomrule
\end{tabular}
\caption{Examples of Question Evaluation Template 1}
\label{tab:at1}
\end{table}

\begin{table}[t]
\small
\centering
\begin{tabular}{@{}>{\centering\arraybackslash}m{0.18\linewidth}>{\centering\arraybackslash}m{0.80\linewidth}@{}}
\toprule
\textbf{Attributes} & \textbf{Templates}\\
\midrule
\textbf{Politeness} & Evaluate the politeness of the following question based on the specified criteria and provide an explanation for your rating.$\backslash$n$\backslash$n$\backslash$n- $\ast$$\ast$1 point - Very Impolite:$\ast$$\ast$ The question is rude or disrespectful, lacking basic courtesy.$\backslash$n- $\ast$$\ast$2 points - Impolite:$\ast$$\ast$ The question may come across as brusque or somewhat disrespectful.$\backslash$n- $\ast$$\ast$3 points - Moderately Polite:$\ast$$\ast$ The question is polite but may lack warmth or additional elements of courteous language.$\backslash$n- $\ast$$\ast$4 points - Quite Polite:$\ast$$\ast$ The question is polite and respectful, using appropriate expressions of courtesy.$\backslash$n- $\ast$$\ast$5 points - Very Polite:$\ast$$\ast$ The question is very polite and courteous, including expressions that show respect and consideration.$\backslash$n$\backslash$n$\backslash$n$\ast$$\ast$Question:$\ast$$\ast$ GIVEN QUESTION$\backslash$n$\backslash$n$\ast$$\ast$Rating:$\ast$$\ast$ \{1-5 points\}$\backslash$n$\backslash$n$\ast$$\ast$Reasoning:$\ast$$\ast$$\backslash$n$\backslash$n$\ast$$\ast$Example:$\ast$$\ast$$\backslash$n$\backslash$n$\ast$$\ast$Question:$\ast$$\ast$ Can you please provide detailed information on the oral medication options available for treating scabies?$\backslash$n$\backslash$n$\ast$$\ast$Rating:$\ast$$\ast$ 5 points$\backslash$n$\backslash$n$\ast$$\ast$Reasoning:$\ast$$\ast$$\backslash$n- The use of "please" shows a high level of politeness and respect, making the question very courteous and considerate in tone.\\\midrule
\textbf{Grammar and Spelling}  & Evaluate the grammar and spelling of the following question based on the specified criteria and provide an explanation for your rating.$\backslash$n$\backslash$n$\backslash$n- $\ast$$\ast$1 point - Very Poor:$\ast$$\ast$ The question contains numerous grammatical and spelling errors.$\backslash$n- $\ast$$\ast$2 points - Poor:$\ast$$\ast$ The question has several grammatical and spelling errors.$\backslash$n- $\ast$$\ast$3 points - Moderate:$\ast$$\ast$ The question has occasional grammatical or spelling errors.$\backslash$n- $\ast$$\ast$4 points - Good:$\ast$$\ast$ The question has minor or no grammatical and spelling errors.$\backslash$n- $\ast$$\ast$5 points - Excellent:$\ast$$\ast$ The question is free from grammatical and spelling errors.$\backslash$n$\backslash$n$\backslash$n$\ast$$\ast$Question:$\ast$$\ast$ GIVEN QUESTION$\backslash$n$\backslash$n$\ast$$\ast$Rating:$\ast$$\ast$ \{1-5 points\}$\backslash$n$\backslash$n$\ast$$\ast$Reasoning:$\ast$$\ast$$\backslash$n$\backslash$n$\ast$$\ast$Example:$\ast$$\ast$$\backslash$n$\backslash$n$\ast$$\ast$Question:$\ast$$\ast$ Can you please provide detailed information on the oral medication options available for treating scabies?$\backslash$n$\backslash$n$\ast$$\ast$Rating:$\ast$$\ast$ 5 points$\backslash$n$\backslash$n$\ast$$\ast$Reasoning:$\ast$$\ast$$\backslash$n- The question is free from grammatical and spelling errors, demonstrating excellent use of language. \\\midrule
\textbf{Structure}  & Evaluate the structure of the following question based on the specified criteria and provide an explanation for your rating.$\backslash$n$\backslash$n$\backslash$n- $\ast$$\ast$1 point - Very Poorly Structured:$\ast$$\ast$ The question is poorly structured and difficult to follow, lacking logical sequence.$\backslash$n- $\ast$$\ast$2 points - Poorly Structured:$\ast$$\ast$ The question has some structure, but it lacks effective organization and logical flow.$\backslash$n- $\ast$$\ast$3 points - Moderately Well-Structured:$\ast$$\ast$ The question is reasonably well-structured but could be improved for better understanding and flow.$\backslash$n- $\ast$$\ast$4 points - Well-Structured:$\ast$$\ast$ The question is well-organized, making it easy to follow and understand.$\backslash$n- $\ast$$\ast$5 points - Excellently Structured:$\ast$$\ast$ The question is excellently structured, enhancing quick and clear comprehension with a logical and coherent sequence.$\backslash$n$\backslash$n$\backslash$n$\ast$$\ast$Question:$\ast$$\ast$ GIVEN QUESTION$\backslash$n$\backslash$n$\ast$$\ast$Rating:$\ast$$\ast$ \{1-5 points\}$\backslash$n$\backslash$n$\ast$$\ast$Reasoning:$\ast$$\ast$$\backslash$n$\backslash$n$\ast$$\ast$Example:$\ast$$\ast$$\backslash$n$\backslash$n$\ast$$\ast$Question:$\ast$$\ast$ Can you please provide detailed information on the oral medication options available for treating scabies?$\backslash$n$\backslash$n$\ast$$\ast$Rating:$\ast$$\ast$ 5 points$\backslash$n$\backslash$n$\ast$$\ast$Reasoning:$\ast$$\ast$$\backslash$n- The question is excellently structured, presenting a clear and direct inquiry. The logical progression from a general request for information to specific details about oral medication options makes it easy for the responder to understand and address. \\
\bottomrule
\end{tabular}
\caption{Examples of Question Evaluation Template 2}
\label{tab:at2}
\end{table}

\begin{table}[t]
\small
\centering
\begin{tabular}{@{}>{\centering\arraybackslash}m{0.18\linewidth}>{\centering\arraybackslash}m{0.80\linewidth}@{}}
\toprule
\textbf{Attributes} & \textbf{Templates}\\
\midrule
\textbf{Neutrality} & Evaluate the neutrality of the following question based on the specified criteria and provide an explanation for your rating.$\backslash$n$\backslash$n$\backslash$n- $\ast$$\ast$1 point - Very Biased:$\ast$$\ast$ The question is heavily biased, containing strong subjective opinions and positions.$\backslash$n- $\ast$$\ast$2 points - Biased:$\ast$$\ast$ The question shows some bias, indicating a leaning towards a particular view or opinion.$\backslash$n- $\ast$$\ast$3 points - Moderately Neutral:$\ast$$\ast$ The question is generally neutral but includes minor subjective elements.$\backslash$n- $\ast$$\ast$4 points - Quite Neutral:$\ast$$\ast$ The question is largely neutral, with very little subjective language.$\backslash$n- $\ast$$\ast$5 points - Completely Neutral:$\ast$$\ast$ The question is entirely neutral, free of any subjective opinions or biases.$\backslash$n$\backslash$n$\backslash$n$\ast$$\ast$Question:$\ast$$\ast$ GIVEN QUESTION$\backslash$n$\backslash$n$\ast$$\ast$Rating:$\ast$$\ast$ \{1-5 points\}$\backslash$n$\backslash$n$\ast$$\ast$Reasoning:$\ast$$\ast$$\backslash$n$\backslash$n$\ast$$\ast$Example:$\ast$$\ast$$\backslash$n$\backslash$n$\ast$$\ast$Question:$\ast$$\ast$ Can you please provide detailed information on the oral medication options available for treating scabies?$\backslash$n$\backslash$n$\ast$$\ast$Rating:$\ast$$\ast$ 5 points$\backslash$n$\backslash$n$\ast$$\ast$Reasoning:$\ast$$\ast$$\backslash$n- The question is presented without any subjective bias, focusing solely on gathering factual information about oral medication options for treating scabies, making it completely neutral.\\\midrule
\textbf{Emotion}  & Evaluate the emotional tone of the following question based on the specified criteria and provide an explanation for your rating.$\backslash$n$\backslash$n$\backslash$n- $\ast$$\ast$1 point - Highly Emotional:$\ast$$\ast$ The question uses highly emotional, exaggerative, or sensational language.$\backslash$n- $\ast$$\ast$2 points - Emotional:$\ast$$\ast$ The question displays some emotional language, though not excessive.$\backslash$n- $\ast$$\ast$3 points - Moderately Emotional:$\ast$$\ast$ The question contains a balance of emotional and neutral language.$\backslash$n- $\ast$$\ast$4 points - Minimally Emotional:$\ast$$\ast$ The question uses minimal emotional language, maintaining an objective tone.$\backslash$n- $\ast$$\ast$5 points - Emotionally Neutral:$\ast$$\ast$ The question is emotionally neutral, using precise and factual language.$\backslash$n$\backslash$n$\backslash$n$\ast$$\ast$Question:$\ast$$\ast$ Can you please provide detailed information on the oral medication options available for treating scabies?$\backslash$n$\backslash$n$\ast$$\ast$Rating:$\ast$$\ast$ \{1-5 points\}$\backslash$n$\backslash$n$\ast$$\ast$Reasoning:$\ast$$\ast$$\backslash$n$\backslash$n$\ast$$\ast$Example:$\ast$$\ast$$\backslash$n$\backslash$n$\ast$$\ast$Question:$\ast$$\ast$ GIVEN QUESTION$\backslash$n$\backslash$n$\ast$$\ast$Rating:$\ast$$\ast$ 5 points$\backslash$n$\backslash$n$\ast$$\ast$Reasoning:$\ast$$\ast$$\backslash$n- The question is posed in a straightforward and factual manner, devoid of any emotional language, making it emotionally neutral. \\\midrule
\textbf{Clarity}  & Evaluate the clarity of the following question based on the specified criteria and provide an explanation for your rating.$\backslash$n$\backslash$n$\backslash$n- $\ast$$\ast$1 point - Very Unclear:$\ast$$\ast$ The question is ambiguous and difficult to understand.$\backslash$n- $\ast$$\ast$2 points - Unclear:$\ast$$\ast$ The question is partially clear but could be misunderstood.$\backslash$n- $\ast$$\ast$3 points - Moderately Clear:$\ast$$\ast$ The question is understandable but could benefit from further clarification.$\backslash$n- $\ast$$\ast$4 points - Quite Clear:$\ast$$\ast$ The question is clear and easy to understand.$\backslash$n- $\ast$$\ast$5 points - Very Clear:$\ast$$\ast$ The question is completely clear, with no ambiguity.$\backslash$n$\backslash$n$\backslash$n$\ast$$\ast$Question:$\ast$$\ast$ Can you please provide detailed information on the oral medication options available for treating scabies?$\backslash$n$\backslash$n$\ast$$\ast$Rating:$\ast$$\ast$ \{1-5 points\}$\backslash$n$\backslash$n$\ast$$\ast$Reasoning:$\ast$$\ast$$\backslash$n$\backslash$n$\ast$$\ast$Example:$\ast$$\ast$$\backslash$n$\backslash$n$\ast$$\ast$Question:$\ast$$\ast$ GIVEN QUESTION$\backslash$n$\backslash$n$\ast$$\ast$Rating:$\ast$$\ast$ 5 points$\backslash$n$\backslash$n$\ast$$\ast$Reasoning:$\ast$$\ast$$\backslash$n- The question is structured and phrased clearly, making it easy to understand and directly addressing the need for information on oral medication options. \\
\bottomrule
\end{tabular}
\caption{Examples of Question Evaluation Template 3}
\label{tab:at3}
\end{table}